# Ratio law: mathematical descriptions for a universal relationship between AI performance and input samples

Boming Kang[2], Qinghua Cui[1,2*]

[1]School of Sports Medicine, Wuhan Institute of Physical Education, No. 461 Luoyu Rd. Wuchang District, Wuhan 430079, Hubei Province, China

[2]Department of Biomedical Informatics, State Key Laboratory of Vascular Homeostasis and Remodeling, School of Basic Medical Sciences, Peking University, 38 Xueyuan Rd, Beijing, 100191, China

* To whom the correspondence should be addressed:

Qinghua Cui, email: cuiqinghua@bjmu.edu.cn

**Abstract**

Artificial intelligence (AI) based on machine learning and deep learning has made significant advances in various fields such as protein structure prediction and climate modeling. However, a central challenge remains: the "black box" nature of AI, where precise quantitative relationships between inputs and outputs are often lacking. Here, by analyzing 323 AI models trained to predict human protein essentiality, we uncovered a ratio law showing that model performance, as measured by the F1 score and the area under the precision-recall curve (AUPRC) can be precisely given by the following two equations: $F1 = \alpha \times r$ and $AUPRC = \beta \times r$, where $r$ represents the ratio of minority to majority samples, $\alpha$ and $\beta$ are dataset-dependent constants. Moreover, we mathematically proved that one AI model will achieve its optimal performance when $r = 1$, i.e., on the balanced dataset. More importantly, we next explore whether this finding can further guide us to enhance AI models' performance. Therefore, we divided the imbalanced dataset into several balanced ($r = 1$) subsets to train base classifiers, and then applied a bagging-based ensemble learning strategy to combine these base models. As a result, the equations-guided strategy substantially improved model performance, with an increase of 4.06% in AUPRC and 5.28% in F1, respectively, which outperformed traditional dataset balancing techniques as well. Finally, we confirmed the broad applicability and generalization of the equations using different types of classifiers and 10 additional and diverse binary classification tasks. In summary, this study reveals two equations precisely linking AI's input and output, which could be helpful for unboxing the mysterious black box of AI.

# 1 Introduction

Artificial intelligence (AI), particularly machine learning and deep learning, has revolutionized various fields. In healthcare, AI systems have improved diagnostic accuracy in medical imaging, with models like convolutional neural networks (CNN) achieving results comparable to human experts[1]. AI has also significantly advanced protein structure prediction, exemplified by DeepMind's AlphaFold, which has transformed structural biology[2]. In climate science, AI models are used for predicting extreme weather and analyzing climate change[3]. Beyond these areas, large language models (LLMs), such as OpenAI's GPT-4, have achieved state-of-the-art (SOTA) results in diverse natural language processing tasks, greatly facilitating our daily life and improving productivity. However, one bottleneck of AI models is that they function as black boxes, offering limited interpretability, especially in terms of precise quantitative formulas linking inputs and outputs. To solve this, OpenAI introduced a scaling law for LLMs, which posits that model performance depends strongly on three factors: the number of model parameters $N$, the size of the dataset $D$, and the amount of compute $C$ used for training. Specifically, model performance follows a power-law relationship with each of these three factors $N$, $D$, and $C$ when not bottlenecked by the others[4]. The scaling law provides three empirical mathematical formulas ($L = (C_{min}/2.3 * 10^8)^{-0.05}$, $L = (D/5.4 * 10^{13})^{-0.095}$, and $L = (N/8.8 * 10^{13})^{-0.076}$) for describing the relationship between model performance (Denoted by L, where L is the test loss for evaluating model performance.) and these influencing factors, enabling the optimal allocation of a fixed compute budget and the estimation of model performance based on $N$, $D$ and $C$.

However, the scaling law does not account for the ratio of minority to majority class samples within a dataset, a key factor that can significantly impact model performance in classification tasks. In reality, some events are common while others are exceedingly rare, such as cancer screenings. As a result, AI models are routinely trained and deployed on class-imbalanced data where relatively few samples are associated with certain minority classes, while majority classes dominate the datasets. As a result, class-imbalanced training datasets often negatively impact model performance[5]. However, there is currently a lack of a precise mathematical formulation to describe how dataset imbalance affects model performance, making it difficult to quantitatively analyze the relationship between model performance and the degree of dataset imbalance.

Essential genes are key components of the minimal genome required for an organism's survival, making them indispensable[6]. Essential proteins encoded by these genes often perform core biological functions related to growth and development[7]. It has been reported that human protein essentiality is context dependent and closely related to cell type and cellular physiological stage[8]. This implies that the ratio of essential to non-essential genes varies across different human cell lines, with the number of essential genes being far fewer than non-essential genes. This provides us with a valuable large-scale data resource for studying how model performance changes with varying minority-to-majority sample ratios in datasets.

Here, we collected a total of 323 binary classification tasks from the Project Score database[9], aiming to distinguish essential genes (minority class) from non-essential genes (majority class) across 323 human cell lines. By training a deep learning model for each human cell line[10], we uncovered a clear mathematical relationship between model performance (measured by F1 score and AUPRC) and the ratio ($r$) of minority to majority samples, expressed as $F1 = \alpha \times r$ and $AUPRC = \beta \times r$, where $\alpha$ and $\beta$ are dataset-dependent constants (**Fig. 1**). It can be mathematically proved that model performance will be peaked when $r = 1$, which further triggers us to explore whether these equations can guide us to enhance AI performance. For doing so, we divided the imbalanced dataset into several balanced subsets to train base classifiers, and then combine them via a bagging-based ensemble learning strategy. We not only demonstrated that the ensemble learning method substantially improved model performance and outperformed traditional dataset balancing techniques on the 323 human cell line datasets, but also validated these findings with 10 additional binary classification tasks. Additionally, we confirmed the broad applicability and generalization of the two equations using different types of classifiers and various binary classification tasks. Overall, the two AI equations we discovered in this study can link the AI models' inputs and outputs, enhancing their interpretability and providing guidance for AI model design.

## 2 Materials and Methods

### 2.1 Dataset Construction

We mainly followed the same methodology described in our previous work[10]. In addition, we calculated the number of essential genes (minority class samples) in each cell line and found significant variation in the number of essential genes across the 323 human cell lines, ranging from 353 to 2117, which could be useful for exploring the relationship between the model performance and the ratio of minority to majority samples.

### 2.2 Model Architecture

Here, we utilized the PIC model architecture from our previous work[10], which were mainly built by fine-tuning a pre-trained protein language model, ESM-2[11].

### 2.3 Model Detail

For all models, we employed 10-fold cross-validation to comprehensively evaluate model performance. Specifically, the entire dataset was divided into 10 subsets, with one subset used as the validation set and the remaining subsets used as the training set in each fold iteration. We followed the optimal hyperparameter configurations consistent with our previous work[10] to ensure that each model was fully trained. All PIC models in this study were retrained based on Python 3.10.13 and implemented in the Pytorch 1.12.1 library. The training processes were executed on a single Nvidia A100 80GB GPU.

### 2.4 Evaluation Metrics

The formulas for the performance evaluation metrics used in this study are as follows:

1. **Accuracy:**

$$Accuracy = \frac{TP + TN}{TP + TN + FP + FN} \tag{1}$$

2. **Precision:**

$$Precision = \frac{TP}{TP + FP} \tag{2}$$

3. **Recall (True Positive Rate, TPR):**

$$TPR = \frac{TP}{TP + FN} \tag{3}$$

4. **False Positive Rate (FPR):**

$$FPR = \frac{FP}{FP + TN} \tag{4}$$

5. **F1 Score:**

$$F1 = 2 \times \frac{Precision \times Recall}{Precision + Recall} \tag{5}$$

6. **Area Under the Receiver Operating Characteristic Curve (AUROC):**

$$AUROC = \int_0^1 TPR(FPR) dFPR \tag{6}$$

7. **Area Under the Precision-Recall Curve (AUPRC):**

$$AUPRC = \int_0^1 Precision(Recall) dRecall \tag{7}$$

**Notation:**

- $TP$ (True Positives): The number of positive instances correctly predicted as positive.

- $TN$ (True Negatives): The number of negative instances correctly predicted as negative.

- $FP$ (False Positives): The number of negative instances incorrectly predicted as positive.

- $FN$ (False Negatives): The number of positive instances incorrectly predicted as negative.

**2.5 Dataset Balancing Methods**

In this study, we utilized various dataset balancing techniques, including undersampling, oversampling, SMOTE (Synthetic Minority Over-sampling Technique) [12], and bagging-based ensemble[13], as detailed below:

**2.5.1 Problem Formulation**

Consider a dataset $D = \{(x_i, y_i)\}_{i=1}^n$, where $x_i$ is the feature vector and $y_i \in \{0,1\}$ denotes the binary class label, with $y = 1$ representing the minority class and $y = 0$ representing the majority class, $n$ is the sample number.

Define:

- $N_0 = \sum_{i=1}^n 1(y_i = 0)$ be the number of majority class samples

- $N_1 = \sum_{i=1}^{n} 1(y_i = 1)$ be the number of minority class samples, where $N_1 < N_0$.

Dataset balancing techniques are used to transform an imbalanced dataset $D$ into a balanced dataset $D'$.

### 2.5.2 Undersampling

The undersampling approach involves randomly selecting a subset of the majority class, $S_0 \subseteq \{i \mid y_i = 0\}$, such that the size of the subset equals the number of minority class samples, i.e., $|S_0| = N_1$.

The resulting balanced dataset $D'$ is defined as:

$$D' = \{(x_i, y_i) \mid i \in S_0\} \cup \{(x_i, y_i) \mid y_i = 1\} \tag{8}$$

Thus, $D'$ contains an equal number of samples from both the majority and minority classes.

### 2.5.3 Oversampling

The oversampling method involves replicating or generating new minority class samples until the number of minority class samples matches the number of majority class samples. This can be achieved by randomly selecting a subset $S_1 \subseteq \{i \mid y_i = 1\}$ and generating additional samples to ensure $|S_1| = N_0$.

The balanced dataset $D'$ is then formed as:

$$D' = \{(x_i, y_i) \mid y_i = 0\} \cup \{(x'_i, y'_i) \mid i' \in S_1\} \tag{9}$$

Where:

- $x'_i$ represents either replicated or synthetically generated minority class samples.
- $D'$ contains an equal number of majority and minority class samples, i.e., $|S_1| = N_0$.

### 2.5.4 Synthetic minority oversampling technique (SMOTE)

SMOTE generates synthetic samples for the minority class by interpolating between existing minority samples. For each minority class sample $x_i$ where $y_i = 1$, a random minority neighbor $x_i^k$ is selected from the k-nearest neighbors of $x_i$. A synthetic sample $x_i^{synthetic}$ is generated as:

$$x_i^{synthetic} = x_i + \lambda(x_i^k - x_i) \tag{10}$$

Where:

- $x_i$ is a randomly selected minority class sample.

- $x_i^k$ is a randomly selected neighbor of $x_i$ from its k-nearest neighbors.
- $\lambda \in [0,1]$ is a random number that determines the interpolation between $x_i$ and $x_i^k$.

The balanced dataset $D'$ is then composed of all majority class samples and the original minority class samples, along with the newly generated synthetic samples:

$$D' = \{(x_i, y_i) \mid y_i = 0\} \cup \{(x_i, y_i) \mid y_i = 1\} \cup \{(x_i^{synthetic}, 1) \mid i = 1, \dots, N_{synthetic}\} \quad (11)$$

Where:

$N_{synthetic}$ is the number of synthetic samples generated to balanced the dataset, such that the final number of minority class samples matches the number of majority class samples.

### 2.5.5 Bagging-based ensemble

**Sampling without replacement (Ensemble-1)**

In this method, each balanced subset $D_k$ is created by randomly selecting without replacement from the majority class. The construction of the k-th dataset can be expressed as:

$$D_k = \{(x_i, y_i) \mid y_i = 1\} \cup \{(x_j, y_j) \mid y_j = 0, j \in S_k\} \quad (12)$$

Where:

- $S_k \subseteq \{j \mid y_j = 0\}$ is a subset of majority class indices, selected without replacement, ensuring that no majority class sample is reused across different subsets separated from the same dataset.
- $|S_k| = N_1$, meaning each subset contains exactly as many majority class samples as minority class samples.

**Sampling with replacement (Ensemble-2)**

Sampling with replacement allows majority class samples to be reused across different subsets. The balanced dataset $D_k$ is formed as:

$$D_k = \{(x_i, y_i) \mid y_i = 1\} \cup \{(x_j, y_j) \mid y_j = 0, j \in S_k'\} \quad (13)$$

Where:

- $S_k' \subseteq \{j \mid y_j = 0\}$ is a subset of majority class indices, selected with replacement, allowing the same majority class sample to appear in multiple subsets.
- $|S_k'| = N_1$, ensuring the subset is balanced by including $N_1$ samples from the majority class.

**The number of base classifiers used in ensemble**

For the sampling without replacement method, the number of base classifiers is:

$$K = \left[\frac{N_0}{N_1}\right] \tag{14}$$

Where:

- $K$ is the total number of base classifiers.
- $N_0$ is the number of majority class samples.
- $N_1$ is the number of minority class samples.
- $[\cdot]$ is the rounding function that rounds a real number to the nearest integer.

For the sampling with replacement method, it is necessary to ensure that each majority class sample is selected for training. Let $P$ represents the probability that a particular majority class sample is not selected in the training data. By adjusting the value of $K$, we can control the value of $P$ to be less than a threshold $\theta$. Therefore, we derive the following equation:

$$P = \left(1 - \frac{N_1}{N_0}\right)^K < \theta \tag{15}$$

then

$$K < \frac{log(\theta)}{log\left(1 - \frac{N_1}{N_0}\right)} \tag{16}$$

Where:

- $\theta$ is a threshold, default set to 0.05.
- $N_0$ is the number of majority class samples.
- $N_1$ is the number of minority class samples.
- $K$ is the total number of base classifiers.

**Hard voting (majority voting) in ensemble**

The hard voting method predicts the final label based on the majority vote across all classifiers. The formula is:

$$y_{pred} = majority\_vote\{f_k(x)\}_{k=1}^{K} \tag{17}$$

Where:

- $y_{pred}$ is the final predicted label, either 0 or 1, determined by the majority vote among all classifiers.

- $majority\_vote\{...\}$ denotes the operation of taking the majority vote from the set of classifier's predictions.
- $f_k(x)$ is the prediction made by the k-th classifier for the input $x$, where $f_k(x) \in \{0,1\}$.
- $K$ is the total number of base classifiers.

**Adaptive voting threshold**

In the hard voting strategy, the adaptive voting threshold is set based on the ratio of positive to negative samples in the dataset, rather than fixing it at 50%.

The formula for calculating the adaptive voting threshold is as follows:

$$V_{threshold} = 1 - \frac{N_1}{N_0 + N_1} \quad (18)$$

The final ensemble prediction $\hat{Y}$ is:

$$\hat{Y} = \begin{cases} 1, if\ \frac{1}{K}\sum_{i=1}^{K} \hat{y}_i \geq V_{threshold} \\ 0, otherwise \end{cases} \quad (19)$$

Where:
- $N_0$ is the number of majority class samples.
- $N_1$ is the number of minority class samples
- $K$ is the total number of base classifiers
- $\hat{y}_i \in \{0,1\}$ is the prediction of the *i*-th base classifier for a given sample, where 0 represents the negative class and 1 represents the positive class.

**Soft voting (weighted averaging) in ensemble**

In soft voting, the final prediction is determined by averaging the probability predictions of all classifiers. The formula is:

$$P(y = 1 \mid x) = \frac{1}{K}\sum_{k=1}^{K} P_k(y = 1 \mid x) \quad (20)$$

Where:
- $P(y = 1 \mid x)$ is the averaged predicted probability that the sample $x$ belongs to class 1.
- $P_k(y = 1 \mid x)$ is the predicted probability assigned by the k-th classifier that the sample $x$ belongs to class 1.
- $K$ is the total number of base classifiers.

## 2.6 Random classifiers used in the stimulation experiments

In this study, we employed the DummyClassifier from the *scikit-learn* package (version 1.5.1) to validate the broad applicability and generalization of the AI equations. This classifier generates predictions without learning from the training data, allowing us to benchmark against a non-informative model. Two distinct strategies "stratified" and "uniform" were utilized to simulate different random prediction behaviors.

### 2.6.1 Stratified strategy

The stratified strategy generates class predictions based on the distribution of classes in the training data. Let $P_i$ represents the probability of observing class $i$ in the training dataset, where $i \in \{0,1\}$ for binary classification. The classifier generates predictions such that:

$$P(\hat{y} = i) = p_i \quad for \ \ i \in \{0,1\} \tag{21}$$

where:

$$p_i = \frac{n_i}{n_0 + n_i} \tag{22}$$

and $n_i$ is the number of samples belongs to class $i$ in the training data, and $\hat{y}$ is the predicted class. This ensures that the predictions reflect the class distribution of the original dataset.

### 2.6.2 Uniform strategy

In the uniform strategy, the classifier predicts each class with equal probability, regardless of the actual class distribution in the training data. For binary classification, the probability of predicting each class is:

$$P(\hat{y} = i) = 0.5 \quad for \ \ i \in \{0,1\} \tag{23}$$

This implies that:

$$P(\hat{y} = 0) = P(\hat{y} = 1) = 0.5 \tag{24}$$

Therefore, predictions are uniformly distributed across all classes, independent of the actual class frequencies in the training set.

## 3 Results

**3.1 Two equations precisely linking AI model performance and sample size**

Given that there are a number of equations precisely describing natural phenomena such as Newton's gravity equation linking gravitational force and mass, we expect that there could be similar relations in AI. To explore this, we utilized 323 AI models for predicting protein essentiality in human cell lines. For each dataset, we defined the number of minority class samples (the essential proteins) as $S_1$ and that of majority class samples (the non-essential proteins) as $S_2$. We then used $r$ to represent the ratio of minority class to majority class samples ($r = \frac{S_1}{S_2}$), which reflects the degree of imbalance in the dataset. Detailed information about the 323 human cell line datasets is provided in **Supplementary Table 1**. Further, we systematically investigated whether the model performance (Accuracy, AUPRC, AUROC, F1, Precision, and Recall) is correlated with $r$ or not. As a result, we found that F1 (*R = 0.8553, p-value = 1.08e-93*), AUPRC (*R = 0.8423, p-value = 3.54e-88*), and Recall (*R = 0.8662, p-value = 1.03e-98*) exhibit a significantly positive correlation with $r$, while Accuracy (*R = -0.9651, p-value = 6.37e-189*) shows a significantly negative correlation. The correlations between AUROC (*R = 0.4248, p-value = 1.39e-15*) and Precision (*R = 0.1641, p-value = 3.10e-03*) with $r$ are relatively weak (**Fig. 2 & Table 1**).

It is well known that when evaluating model performance on imbalanced binary classification datasets, F1 and AUPRC provide a more accurate reflection of model performance, as they focus more on the model's ability to classify the minority class. As a result, we revealed the following two equations to link F1 with $r$ and AUPRC with $r$, respectively:

$$F1 = \alpha \times r \qquad (Eq. 1)$$

$$AUPRC = \beta \times r \qquad (Eq. 2)$$

Here, $r$ represents the ratio of minority to majority class samples in the dataset, $\alpha$ and $\beta$ are constants independent of the $r$.

For these 323 human cell line datasets, α = 3.49 and β = 2.72, leading to the following two

equations:

$$F1 = 3.49 \times r \quad (Eq.\ 3)$$

$$AUPRC = 2.72 \times r \quad (Eq.\ 4)$$

**3.2 Mathematical derivation of the AI equations**

Next, we used an ideal random classifier to derive the maximum values of F1 and AUPRC, and the corresponding value of $r$ at that point.

Let r denotes the ratio of minority to majority samples in the dataset:

$$r = \frac{N_1}{N_0} \quad (25)$$

where:
- $N_0$ is the number of majority class samples.
- $N_1$ is the number of minority class samples

The proportion of positive (minority) samples $P_p$ and negative (majority) samples $P_n$ are then:

$$P_p = \frac{r}{1+r} \quad (26)$$

$$P_n = \frac{1}{1+r} \quad (27)$$

For a random binary classifier predicting each class with uniform probability 0.5, the expected values in the confusion matrix are:

- True Positives (TP):

$$TP = P_p \times 0.5 \quad (28)$$

- False Positives (TP):

$$FP = P_n \times 0.5 \quad (29)$$

- False Negatives (FN):

$$FN = P_p \times 0.5 \quad (30)$$

- True Negatives (TN):

$$TN = P_n \times 0.5 \quad (31)$$

**Precision($P$)** is defined as:

$$P = \frac{TP}{TP + FP} \quad (32)$$

**Recall($R$)** is defined as:

$$R = \frac{TP}{TP + FN} \tag{33}$$

### 3.2.1 F1 score equation derivation

Substituting the expected values:

$$P = \frac{P_p \times 0.5}{(P_p + P_n) \times 0.5} = \frac{P_p}{P_p + P_n} \tag{34}$$

$$R = \frac{P_p \times 0.5}{P_p \times 0.5 + P_p \times 0.5} = 0.5 \tag{35}$$

Since $P_p + P_n = 1$. Therefore,

$$P = P_p = \frac{r}{1 + r} \tag{36}$$

The F1 score is the harmonic mean of precision and recall:

$$F1 = 2 \times \frac{P \times R}{P + R} \tag{37}$$

Substituting $P$ and $R$:

$$F1 = \frac{2r}{3r + 1} \tag{38}$$

To analyze the extrema of $F1(r) = \frac{2r}{3r+1}$ within the interval $0 < r \leq 1$, we conducted a differential analysis to examine its monotonic behavior.

We first computed the derivative of $F1(r)$ with respect to $r$:

$$F1'(r) = \frac{d}{dr}\left(\frac{2r}{3r + 1}\right) \tag{39}$$

Applying the quotient rules, the derivate is:

$$F1'(r) = \frac{(2) \times (3r + 1) - (2r) \times (3)}{(3r + 1)^2} \tag{40}$$

Simplifying the numerator:

$$F1'(r) = \frac{2}{(3r + 1)^2} \tag{41}$$

Since $(3r + 1)^2 > 0$ for all $r > 0$, it follows that $F1'(r) > 0$ throughout the interval $0 < r \leq 1$. This indicates that $F1'(r)$ is strictly increasing on this interval.

To find the extrema, we evaluated $F1'(r)$ at the endpoints of the interval.

At $r = 1$:

$$F1(1) = \frac{2 \times 1}{3 \times 1 + 1} = \frac{2}{4} = 0.5 \tag{42}$$

As $r$ approaches 0 from the right:

$$\lim_{r \to 0^+} F1(r) = \lim_{r \to 0^+} \frac{2r}{3r + 1} = 0 \tag{43}$$

Therefore, the minimum value of $F1(r)$ on the interval $0 < r \leq 1$ is 0 (as $r \to 0^+$), and the maximum value is 0.5 (at $r = 1$).

**3.2.2 AUPRC equation derivation**

The AUPRC for a random binary classifier can be mathematically derived based on the class proportions.

For a random classifier, the precision at any recall level is equal to the proportion of positive samples in the dataset, $P_p$. Therefore, the precision-recall curve is a horizontal line at $P = P_p$

The, the AUPRC can be described as:

$$AUPRC = \int_0^1 P(R)dR = \int_0^1 P_p(R)dR = P_p \tag{44}$$

Substituting $P_p$:

$$AUPRC = \frac{r}{1 + r} \tag{45}$$

To analyze the extrema of $AUPRC(r) = \frac{r}{r+1}$ within the interval $0 < r \leq 1$, we performed a differential analysis to examine its monotonic behavior.

We first computed the derivative of $AUPRC(r)$ with respect to $r$:

$$AUPRC'(r) = \frac{d}{dr}\left(\frac{r}{1+r}\right) \tag{46}$$

Applying the quotient rule, the derivate is:

$$AUPRC'(r) = \frac{(1) \times (1+r) - (r) \times (1)}{(1+r)^2} \tag{47}$$

Simplifying the numerator:

$$AUPRC'(r) = \frac{1}{(1+r)^2} \tag{48}$$

Since $(1 + r)^2 > 0$ for all $r > 0$, it follows that $AUPRC'(r) > 0$ throughout the interval $0 < r \leq 1$. This indicates that $AUPRC'(r)$ is strictly increasing on this interval.

To find the extrema, we evaluated $AUPRC(r)$ and the endpoints of the interval.

At $r = 1$:

$$AUPRC(1) = \frac{1}{1+1} = \frac{1}{2} = 0.5 \tag{49}$$

As $r$ approaches 0 from the right:

$$\lim_{r \to 0^+} AUPRC(r) = \lim_{r \to 0^+} \frac{r}{1+r} = 0 \tag{50}$$

Therefore, the minimum value of $AUPRC(r)$ on the interval $0 < r \leq 1$ is 0 (as $r \to 0^+$), and the maximum value is 0.5 (at $r = 1$).

**3.3 Training using balanced datasets separated from one unbalanced dataset enhances model performance**

As indicated by the two equations we revealed above, if the total number of samples is fixed, F1 and AUPRC will achieve their maximum values when $r = 1$. This aligns with our prior understanding that training on balanced datasets could improve model performance. Therefore, the revealed AI equations can serve as a guide for enhancing model performance. To verify this, for the 323 AI models, we split each unbalanced dataset into $K$ balanced subsets, where each subset satisfies $r = 1$. We then trained a separate base classifier on each subset, resulting in $K$ base classifiers, which were ultimately combined using an ensemble learning strategy to form the final model. We utilized and compared two different ensemble learning strategies: sampling with replacement and sampling without replacement **(see section 2.5.5 for details)**. To obtain the prediction results of the ensemble model, we employed both hard voting and soft voting methods to generate the predicted labels and probabilities, respectively **(see section 2.5.5 for details)**.

Due to the complexity of the ensemble learning method, we first examined the effect of the voting threshold in the hard voting method on the F1 score. The voting threshold refers to the proportion of base models predicting a positive class required for the final prediction to be positive. The results showed that the F1 score gradually improved as the voting threshold increased, reaching its maximum when the voting threshold was set to 90% (**Fig. 3a**). Additionally, we implemented an adaptive voting threshold, where the threshold was dynamically calculated based on the ratio of positive to negative samples in the dataset **(see section 2.5.5 for details)**. The results indicated that models using the adaptive voting threshold performed slightly better than those with a fixed voting threshold of 90% (**Fig. 3a**). Therefore, we ultimately set a specific adaptive voting threshold for each ensemble model. Next, we investigated the effect of soft voting, comparing two approaches: averaging the predicted probabilities of the base models versus taking the maximum probability as the final prediction for the ensemble model. The results showed that models using the average probability outperformed those using the maximum probability, with significantly higher AUROC (independent sample t-test, *p-value = 2.08e-3*) and AUPRC (independent sample t-test, p-value = *2.71e-8*) (**Fig. 3b & 3c**).

Further, we systematically compared the performance of the models trained using the ensemble learning approach with those trained directly on the unbalanced datasets. Additionally, we compared the ensemble learning approach with other classic dataset balancing methods, including undersampling, oversampling, and SMOTE **(see section 2.5 for details)**. To ensure the reliability of the results, we applied 10-fold cross-validation to all of the 323 models trained using each method and used the average metric on the test datasets as the final performance for each model. We first calculated the average performance of the 323 AI models across each fold and then compared the model performance of different methods using 10-fold cross-validation. The detailed 10-fold cross-validation results for different dataset balancing methods are provided in **Supplementary Table 2**. The results showed that, compared to models trained directly on imbalanced datasets (Unbalanced), the Ensemble-1 method (sampling without replacement) significantly improved AUPRC by (independent sample t-test, *p-value = 1.06e-3*), F1 (independent sample t-test, *p-value = 1.32e-5*), and AUROC (independent sample t-test, *p-value = 2.24e-10*). Additionally, when compared to other dataset balancing methods, including undersampling, oversampling, and SMOTE, the Ensemble-1 method also significantly enhanced AUPRC, F1, and AUROC. Furthermore, we found no significant differences in the improvements of AUPRC, F1, and AUROC between the Ensemble-1 and Ensemble-2 methods (sampling with replacement) (**Fig. 3d-3f**).

Next, we used the average performance of the 323 models across the 10-fold cross-validation as the final performance for each model and conducted paired sample t-tests to compare the model performance across different dataset balancing methods. The average performance of the 323 models in 10-fold cross-validation is presented in **Supplementary Table 3.** As a result, compared to the Unbalanced method, the Ensemble-1 method significantly improved AUPRC by 4.06% (paired sample t-test, *p-value = 7.96e-219*) and F1 by 5.28% (paired sample t-test, *p-value = 3.23e-169*). AUROC also increases by 4.56% (paired sample t-test, *p-value = 2.39e-206*). Additionally, compared to the undersampling method, the Ensemble-1 method significantly improved AUPRC by 8.62% (paired sample t-test, *p-value = 4.72e-285*), F1 by 9.52% (paired t-test sample, *p-value = 3.47e-304*), and AUROC by 3.07% (paired t-test sample, *p-value = 8.67e-289*). Compared to the Oversampling method, the Ensemble-1 method significantly improved AUPRC by 7.15% (paired t-test sample, *p-value = 2.62e-274*), F1 by 3.92% (paired t-test sample, *p-value = 7.0e-163*), and AUROC by 5.96% (paired t-test sample, *p-value = 1.88e-220*). Finally, compared to the SMOTE method, the Ensemble-1 method significantly improved AUPRC by 8.16% (paired t-test sample, *p-value = 3.04e-297*), F1 by 5.57% (paired t-test sample, *p-value = 3.74e-204*), and AUROC by 7.62% (paired t-test sample, *p-value = 1.92e-230*) (**Fig. 3g-3i & Table 2-3**).

### 3.4 Generalization and applicability of the AI equations

To validate the generalization of the discovered AI equations, we first explored whether it could be applied to different types of AI models. Here, we selected four commonly used models as classifiers to predict human protein essentiality, including random forests [14], XGBoost [15], LightGBM [16] and MLP [17]. The results showed that for all four models, both F1 and AUPRC exhibited a significant positive correlation with $r$, consistent with the relationships described in Eq. 1 and Eq. 2 (**Fig. 4a & 4b**). Furthermore, by comparing the average performance of the models trained on 323 datasets, we found that the MLP model significantly outperformed the other three models (**Fig. 4c & 4d**). These findings indicate that the revealed AI equations, suggesting the revealed equations are universal across various types of machine learning models. Next, we explored whether the AI equation is applicable to other types of binary classification tasks. To this end, we collected 10 additional binary classification datasets from previous studies[18], including tasks such as predicting antimicrobial peptides from protein sequences. These datasets exhibit varying degrees of class imbalance. Using the same methodology as for the 323 AI models trained for predicting human protein essentiality. We trained models on these 10 datasets and examined whether the model performance and $r$ still align with the revealed AI equations. As a result, we observed a strong positive correlation between F1, AUPRC, and $r$ for these 10 models, consistent with the relationships described by Eq. 1 and Eq. 2 (**Fig.5a-5b & Table 4**), suggesting the revealed equations are universal across different types of classification tasks.

To further generalize the two equations, we conducted simulation experiments using binary classification datasets generated by the *scikit-learn* package to validate the impact of varying positive-to-negative sample ratios on model performance over a broader range. Here, we constructed a simple logistic regression model and trained it on datasets with different positive (minority)-to-negative (majority) sample ratios (range: 0.01~0.99). Additionally, we set varying classification difficulties for the datasets to simulate different levels of model performance. We also included two random classifiers, unaffected by the training data, to further verify the applicability of the AI equation across models with varying performance (**see section 2.6 for details**). As a result, as the $r$ increases from 0 to 1, both the AUPRC and F1 of the logistic regression model and the two random classifiers exhibit an upward trend. Additionally, for classification tasks of varying difficulty, the performance of all three models improves as the ratio increases, reaching its maximum when $r = 1$ (**Fig. 5c**).

### 3.5 Mathematical principles underlying the AI equations

Since the prediction results of AI models are influenced by the quality of the training data and model parameters, these predictions exhibit considerable randomness, making it difficult to derive the deep mathematical principles underlying the proposed AI equation. Therefore, we assumed an ideal classifier that is unaffected by training data or model parameters and consistently predicts both positive and negative samples with 50% accuracy. Based on the mathematical derivation in section 3.2, we arrived at the following two equations:

$$\text{F1} = \frac{2r}{3r + 1} \quad \quad \text{(Eq. 5)}$$

$$AUPRC = \frac{r}{r + 1} \quad \quad \text{(Eq. 6)}$$

When $r \in (0,1)$, the function plots about these two equations are shown in **Fig.5d and 5e**, respectively, where F1 and AUPRC both increase as $r$ increases. For the 323 datasets used in this study, $r \in (0, 0.15)$, indicating that they are highly imbalanced datasets. If $r$ is very small ($r \ll 1$), the two equations will be approximated as follows:

$$F1 = \frac{2r}{3r + 1} \approx 2r = \alpha \times r \quad \quad \text{(Eq. 7)}$$

$$AUPRC = \frac{r}{r + 1} \approx r = \beta \times r \quad \quad \text{(Eq. 8)}$$

At this point, the two equations take the same form as Eq. 1 and Eq. 2. It should be noted that coefficient $\alpha$ and $\beta$ may vary depending on the dataset and model parameters.

## 4 Discussion and conclusion

In this study, we reveal a novel mathematical equation linking AI model performance, specially F1 score and AUPRC, to the ratio of positive (minority) and negative (majority) samples in a dataset. By analyzing 323 binary classification tasks on human protein essentiality, we found that both F1 and AUPRC exhibit a strong linear correlation with the ratio of positive and negative samples, expressed by $F1 = \alpha \times r$ and $AUPRC = \beta \times r$, where $r$ represents the sample ratio and $\alpha$ and $\beta$ are dataset-dependent constants. These findings provide valuable insights into how dataset composition, particularly in imbalanced classification tasks, affects AI model performance. We first mathematically proved that the model performance will be optimal when $r = 1$, that is, datasets are balanced, and then confirmed that the optimal performance is achieved when datasets are balanced, highlighting the importance of carefully considering sample ratios in model design. Our experiments showed that dividing an imbalanced dataset into balanced subsets and applying ensemble learning substantially improves model performance, outperforming traditional balancing techniques like undersampling, oversampling, and SMOTE. The revealed equations were further validated across different machine learning models and various classification tasks, suggesting that the two equations are universal. Finally, simulation experiments using synthetic datasets also confirmed the broad applicability of these equations, demonstrating their relevance even for random classifiers and models of varying complexity.

While the revealed equations are promising, they were primarily tested on binary classification tasks, and further research is needed to explore their applicability in multi-class and other AI problem settings. Future studies could also investigate the impact of model complexity and data dynamics on these equations to expand their utility across different AI domains.

In summary, this study provides a precise mathematical framework to better understand AI model performance in imbalanced classification tasks, offering a practical tool for improving model design. The findings have broad applicability and contribute to the ongoing effort to make AI more interpretable and reliable across diverse applications.

ACKNOWLEDGEMENTS

This study was supported by the grants from the National Natural Science Foundation of China [62025102].

**Author contributions**

QC presented the original idea. BK performed the study. BK and QC wrote the raw manuscript and edited the manuscript.

**Reference**

1. Esteva, A. *et al.* Dermatologist-level classification of skin cancer with deep neural networks. *Nature* **542**, 115-118 (2017). https://doi.org:10.1038/nature21056
2. Jumper, J. *et al.* Highly accurate protein structure prediction with AlphaFold. *Nature* **596**, 583-589 (2021). https://doi.org:10.1038/s41586-021-03819-2
3. Rasp, S. *et al.* WeatherBench: A Benchmark Data Set for Data-Driven Weather Forecasting. *Journal of Advances in Modeling Earth Systems* **12**, e2020MS002203 (2020). https://doi.org:https://doi.org/10.1029/2020MS002203
4. Kaplan, J. *et al.* Scaling laws for neural language models. *arXiv preprint arXiv:2001.08361* (2020).
5. Shwartz-Ziv, R., Goldblum, M., Li, Y., Bruss, C. B. & Wilson, A. G. Simplifying neural network training under class imbalance. *Advances in Neural Information Processing Systems* **36** (2024).
6. Bartha, I., Di Iulio, J., Venter, J. C. & Telenti, A. Human gene essentiality. *Nature Reviews Genetics* **19**, 51-62 (2018).
7. Ji, X., Rajpal, D. K. & Freudenberg, J. M. The essentiality of drug targets: an analysis of current literature and genomic databases. *Drug Discovery Today* **24**, 544-550 (2019).
8. Li, Y., Zeng, M., Zhang, F., Wu, F.-X. & Li, M. DeepCellEss: cell line-specific essential protein prediction with attention-based interpretable deep learning. *Bioinformatics* **39**, btac779 (2023).
9. Dwane, L. *et al.* Project Score database: a resource for investigating cancer cell dependencies and prioritizing therapeutic targets. *Nucleic Acids Research* **49**, D1365-D1372 (2021).
10. Kang, B., Fan, R., Cui, C. & Cui, Q. Comprehensive prediction and analysis of human protein essentiality based on a pre-trained protein large language model. *bioRxiv*, 2024.2003. 2026.586900 (2024).
11. Lin, Z. *et al.* Evolutionary-scale prediction of atomic-level protein structure with a language model. *Science* **379**, 1123-1130 (2023).

12	Chawla, N. V., Bowyer, K. W., Hall, L. O. & Kegelmeyer, W. P. SMOTE: synthetic minority over-sampling technique. *Journal of artificial intelligence research* **16**, 321-357 (2002).
13	Breiman, L. Bagging predictors. *Machine learning* **24**, 123-140 (1996).
14	Breiman, L. Random forests. *Machine learning* **45**, 5-32 (2001).
15	Chen, T. & Guestrin, C. in *Proceedings of the 22nd acm sigkdd international conference on knowledge discovery and data mining.*   785-794.
16	Ke, G. *et al.* Lightgbm: A highly efficient gradient boosting decision tree. *Advances in neural information processing systems* **30** (2017).
17	Rumelhart, D. E., Hinton, G. E. & Williams, R. J. Learning representations by back-propagating errors. *nature* **323**, 533-536 (1986).
18	Medina-Ortiz, D. *et al.* Protein Language Models and Machine Learning Facilitate the Identification of Antimicrobial Peptides. *International Journal of Molecular Sciences* **25**, 8851 (2024).

**Figure**

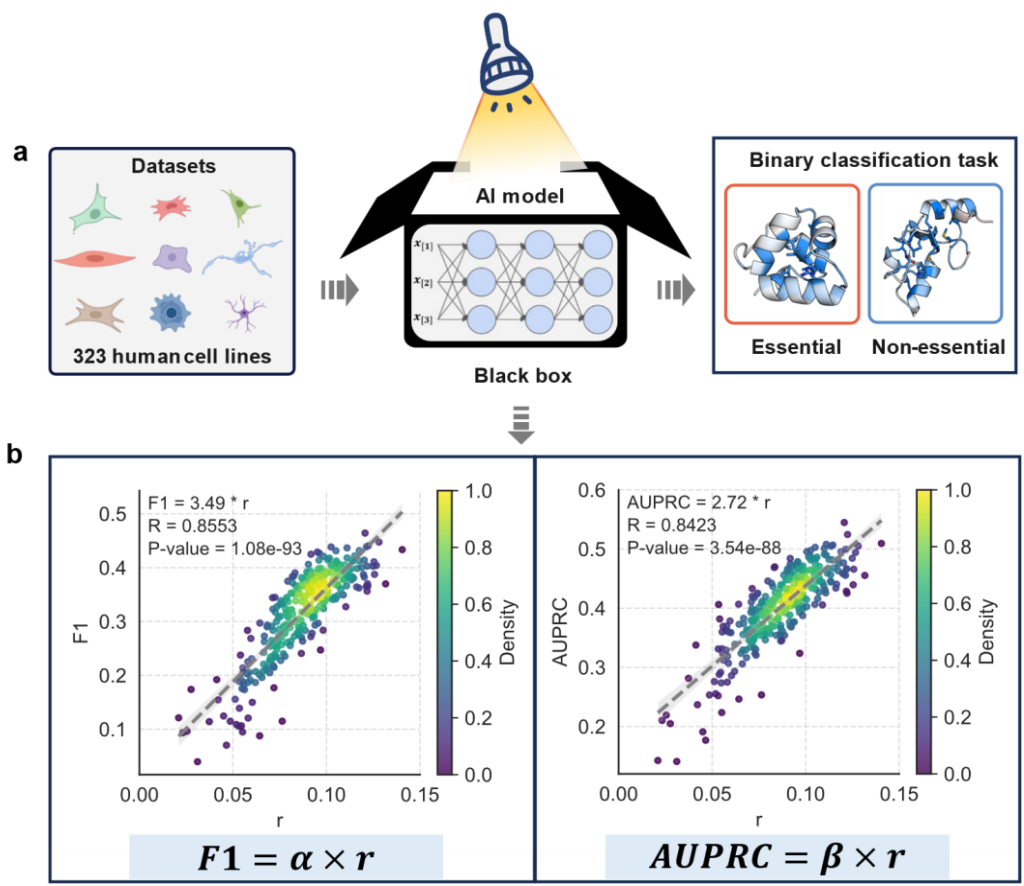

**Fig. 1: Overview of the revealed AI equations. (a)** A total of 323 binary classification datasets are collected for predicting human protein essentiality across human cell lines (left); A specific deep learning model is trained for each dataset, a black-box approach, lacking model interpretability (middle). The task is to distinguish essential proteins from non-essential proteins across 323 human cell lines, with each cell line exhibiting a varying ratio ($r$) (right). **(b)** Two AI equations are uncovered to describe the relationship between the F1 score and the $r$, as well as the relationship between the AUPRC and the $r$, linking the model's inputs and outputs. AUPRC: Area Under the Precision-Recall Curve; $r$: the ratio of minority to majority class samples in the datasets; $\alpha$: a dataset-dependent constant of F1 score. $\beta$: a dataset-dependent constant of AUPRC. R: Pearson correlation coefficient.

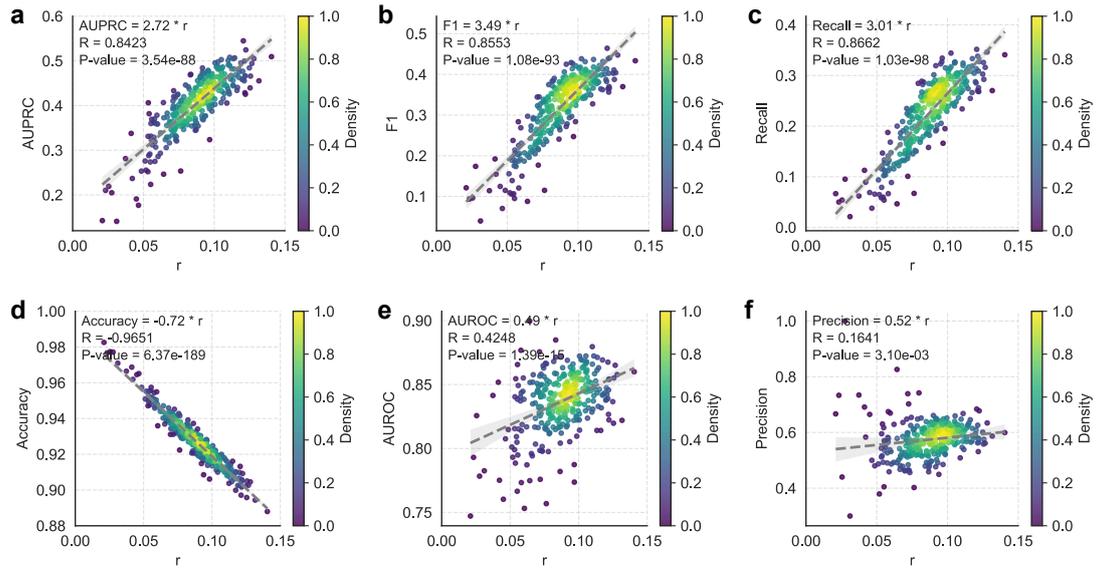

**Fig. 2: The relationship between the model performance and the ratio (*r*).** AUPRC: Area Under the Precision-Recall Curve; AUROC: Area Under the Receiver Operating Characteristic Curve; *r*: the ratio of minority to majority class samples in the dataset. R: Pearson correlation coefficient. The dashed line in each plot is generated using linear fitting. **(a):** AUPRC **(b):** F1 **(c):** Recall **(d):** Accuracy **(e):** AUROC **(f):** Precision.

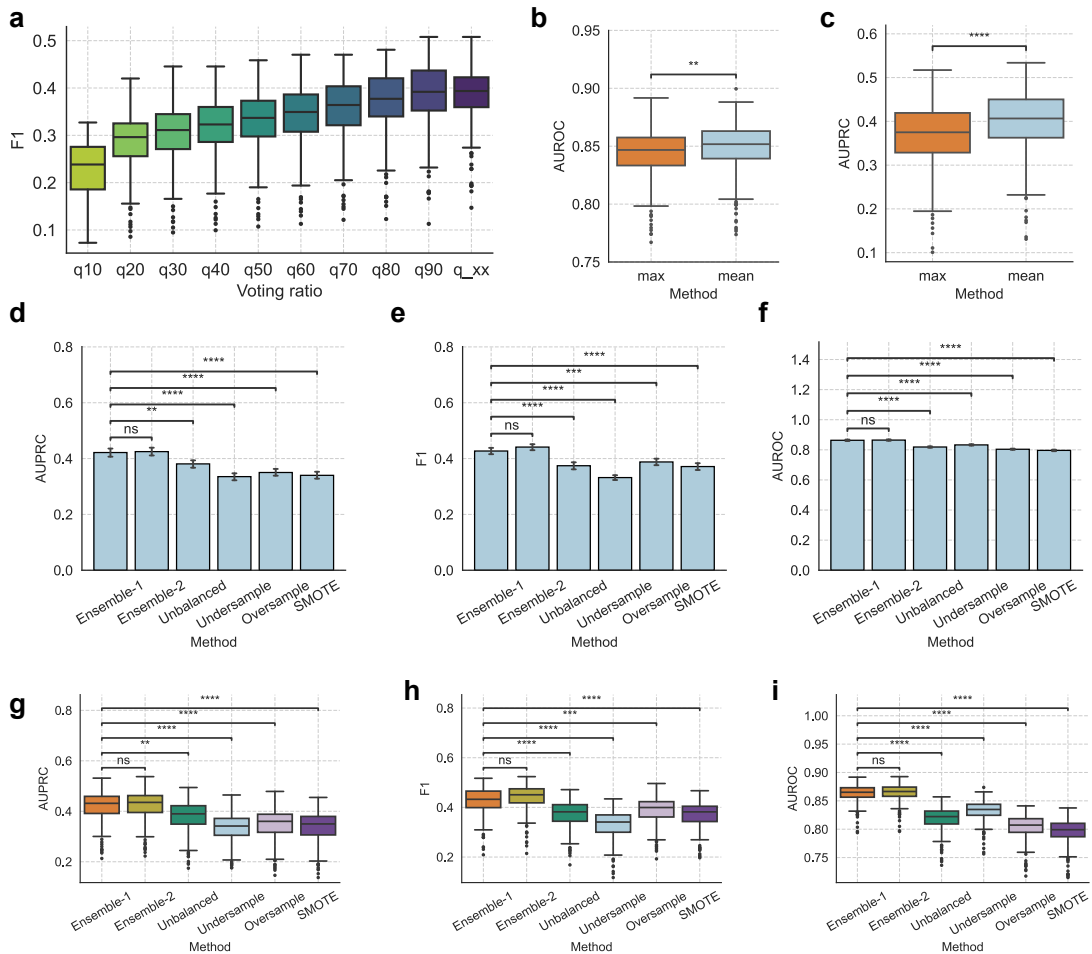

**Fig. 3: Evaluation of model performance using different dataset balancing techniques.** AUPRC: Area Under the Precision-Recall Curve; AUROC: Area Under the Receiver Operating Characteristic Curve; Ensemble-1: the bagging-based ensemble method that sampling without replacement; Ensemble-2: the bagging-based ensemble method that sampling with replacement; Unbalanced: models are trained without using any dataset balancing methods. Undersample: the undersampling method. Oversample: the oversampling method. SMOTE: the synthetic minority over-sampling technique. **(a)** The correlation between the F1 score of the ensemble models and the voting thresholds. q10 to q90 represent voting thresholds ranging from 10% to 90% and q_xx refers to the adaptive voting threshold; **(b-c)** The comparison of the performance of ensemble models using different soft voting approaches. max represents the approach that taking the maximum probability as the final prediction for the ensemble model; mean represents the approach that averaging the predicted probabilities of the base classifiers. **(d-f)**: The comparison of the performance of the ensemble method and other dataset balancing techniques in the 10-fold cross-validation. Statistical significance was determined using independent sample t-tests. ns refers to p-value>0.05; ** refers to p-value<0.01; **** refers to p-value<0.0001. **(g-i):** The paired performance comparison of 323 models using different dataset balancing techniques. Statistical significance was determined using paired sample t-tests. ns refers to p-value>0.05; ** refers to p-value<0.01; **** refers to p-value<0.0001.

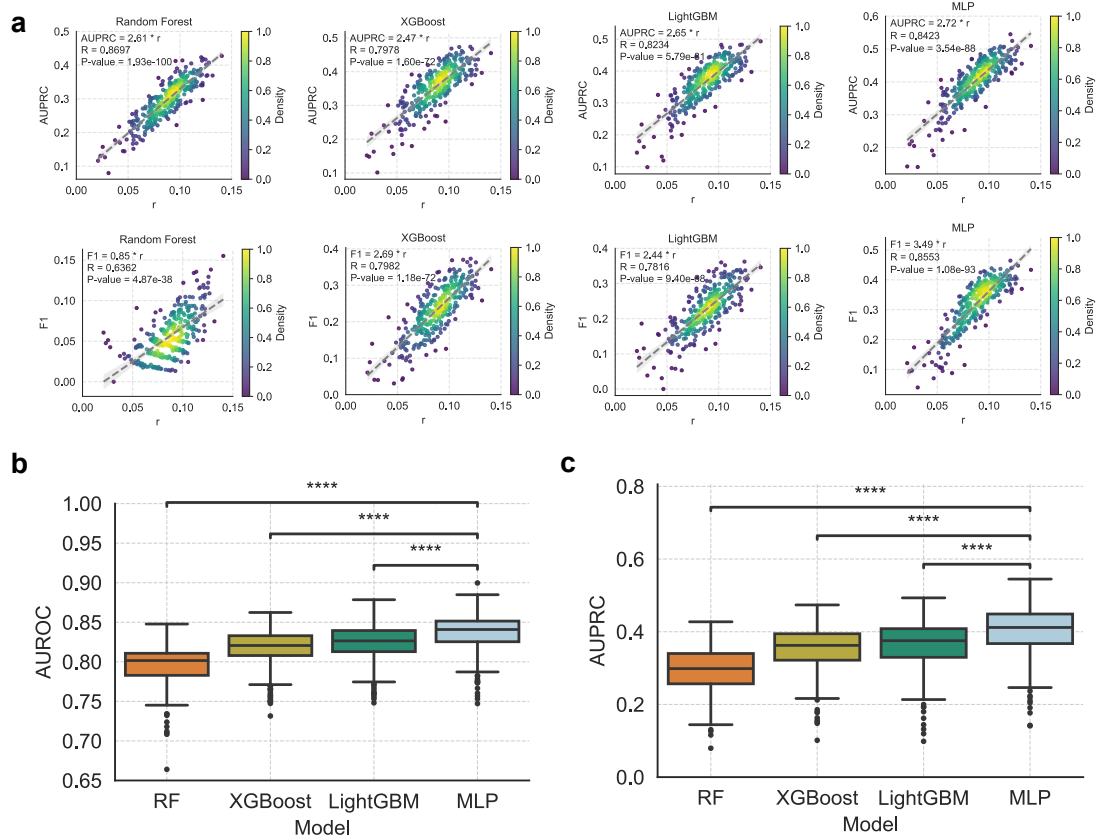

**Fig. 4: The applicability of the AI equations across different types of AI models.** AUPRC: Area Under the Precision-Recall Curve; AUROC: Area Under the Receiver Operating Characteristic Curve; *r*: the ratio of positive to negative samples in the dataset; R: Pearson correlation coefficient. RF: Random Forests. **(a)** The correlations between AUPRC and *r* for the four types of AI models: Random Forests, XGBoost, LightGBM, and MLP. **(b)** The correlations between F1 and *r* for the four types of AI models: Random Forests, XGBoost, LightGBM, and MLP. **(c)** Performance comparison of AUROC for the four AI models across 323 datasets. **(d)** Performance comparison of AUPRC for the four AI models across 323 datasets. Statistical significance was determined using independent sample t-tests. **** refers to p-value<0.0001.

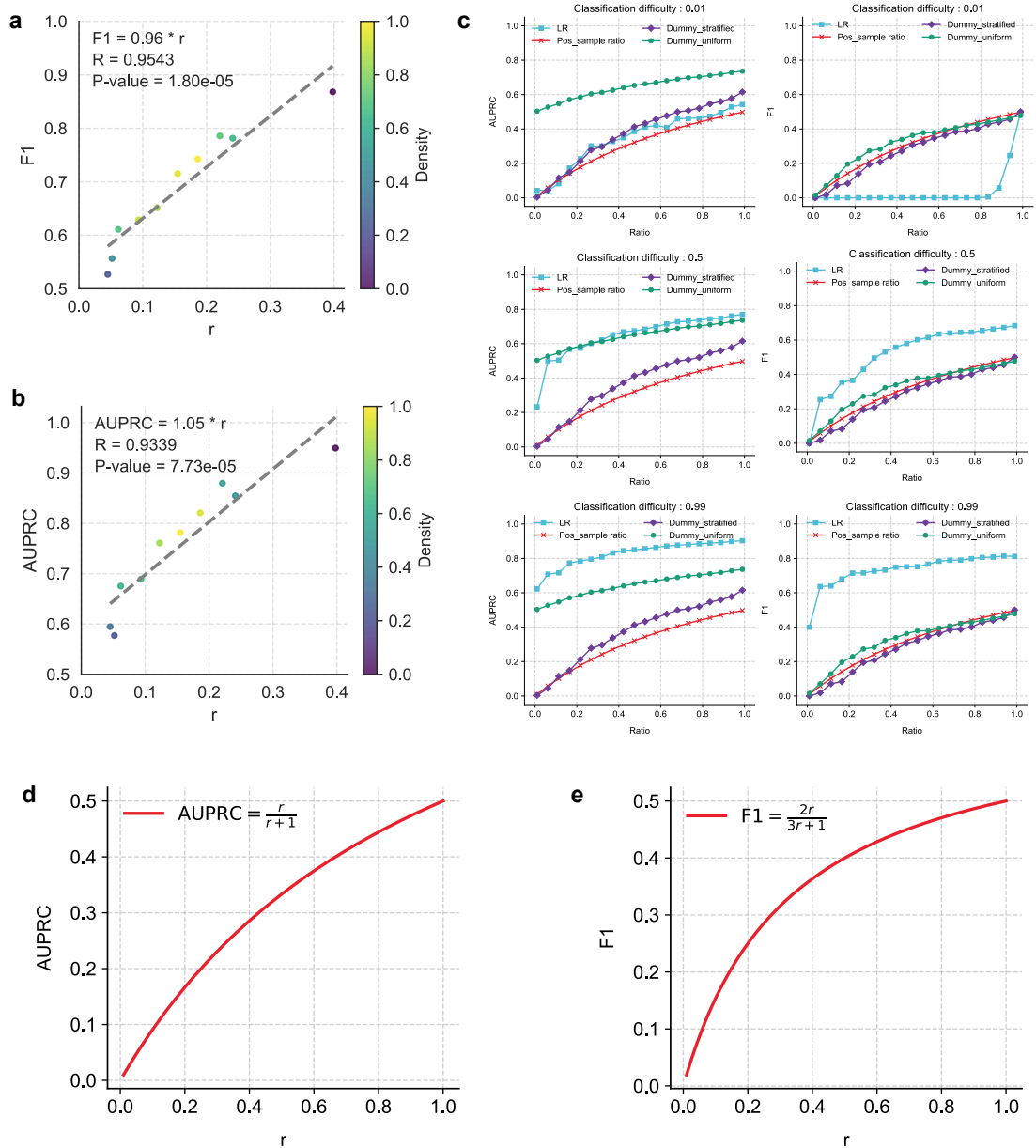

**Fig. 5: The applicability of the AI equations across different types of binary classification tasks.**
AUPRC: Area Under the Precision-Recall Curve; *r*: the ratio of minority to majority class samples in the dataset. R: Pearson correlation coefficient. **(a)** The correlation between F1 and *r* on 10 additional binary classification datasets. **(b)** The correlation between AUPRC and *r* on 10 additional binary classification datasets. **(c)** The stimulation experiments using binary classification datasets generated by *scikit-learn* package. LR: logistic regression model; Dummy_stratified: the random classifier using the stratified strategy; Dummy_uniform: the random classifier using the uniform strategy. Pos_sample ratio: the proportion of positive samples in the dataset. **(d)** The function graph of AUPRC versus *r* for an ideal random classifier. **(e)** The function graph of F1 versus *r* for an ideal random classifier.

**Table**

**Table 1 Correlation analysis results between model performance metrics and *r*.**

| Metric | R | P-value |
| --- | --- | --- |
| AUPRC | 0.8423 | 3.54E-88 |
| F1 | 0.8553 | 1.08E-93 |
| Recall | 0.8662 | 1.03E-98 |
| Accuracy | -0.9651 | 6.37E-89 |
| AUROC | 0.4248 | 1.39E-15 |
| Precision | 0.1641 | 3.10E-03 |

**Note:**

R: Pearson correlation coefficient.

**Table 2 Paired t-test results for model performance using different dataset balancing methods.**

| Metric | Method-1 | Method-2 | Statistics | P-value |
| --- | --- | --- | --- | --- |
| AUPRC | Ensemble-1 | Unbalanced | 82.62 | 7.96E-219 |
| AUPRC | Ensemble-1 | Undersample | 134.56 | 4.72E-285 |
| AUPRC | Ensemble-1 | Oversample | 124.42 | 2.62E-274 |
| AUPRC | Ensemble-1 | SMOTE | 147.02 | 3.04-297 |
| F1 | Ensemble-1 | Unbalanced | 56.52 | 3.23E-169 |
| F1 | Ensemble-1 | Undersample | 154.62 | 3.47E-304 |
| F1 | Ensemble-1 | Oversample | 53.76 | 7E-163 |
| F1 | Ensemble-1 | SMOTE | 73.98 | 3.74E-204 |
| AUROC | Ensemble-1 | Unbalanced | 75.22 | 2.39E-206 |
| AUROC | Ensemble-1 | Undersample | 138.27 | 8.67E-289 |
| AUROC | Ensemble-1 | Oversample | 83.63 | 1.88E-220 |
| AUROC | Ensemble-1 | SMOTE | 90.1 | 1.92E-230 |

**Table 3 Average model performance of different dataset balancing methods in 10-fold cross-validation.**

| Method | AUPRC | AUROC | F1 | Recall | Precision | Accuracy |
| --- | --- | --- | --- | --- | --- | --- |
| Ensemble-1 | 0.4216 | 0.8634 | 0.4271 | 0.6105 | 0.3331 | 0.8726 |
| Ensemble-2 | 0.4249 | 0.8647 | 0.4412 | 0.5578 | 0.3687 | 0.8903 |
| Oversample | 0.3501 | 0.8038 | 0.3879 | 0.3960 | 0.3872 | 0.9027 |
| SMOTE | 0.3400 | 0.7962 | 0.3714 | 0.3617 | 0.3896 | 0.9048 |
| Unbalanced | 0.3810 | 0.8187 | 0.3743 | 0.3142 | 0.4818 | 0.9188 |
| Undersample | 0.3354 | 0.8327 | 0.3319 | 0.7561 | 0.2154 | 0.7602 |

**Table 4 Model performance of different dataset balancing methods on 10 additional binary classification tasks.**

| Task | F1 | AUPRC | r |
| --- | --- | --- | --- |
| Antimicrobial | 0.8681 | 0.9494 | 0.3979 |
| Antibacterial | 0.7814 | 0.8547 | 0.2411 |
| Toxic | 0.7859 | 0.8797 | 0.2209 |
| Anti_gram_pos | 0.7426 | 0.8208 | 0.1862 |
| Anti_gram_neg | 0.7152 | 0.7814 | 0.1547 |
| Metabolic | 0.6513 | 0.7608 | 0.1225 |
| Anti_mammalian_cell | 0.6287 | 0.6892 | 0.0934 |
| Neuropeptide | 0.6109 | 0.6753 | 0.0617 |
| Immunological | 0.5565 | 0.5771 | 0.0518 |
| Anti_inflamatory | 0.5268 | 0.5946 | 0.0451 |